\documentclass[11pt]{article} 
\usepackage{rldmsubmit,palatino}
\usepackage{graphicx}

\usepackage{amsfonts, amsmath, amssymb}
\usepackage{color}
\usepackage{algorithm}
\usepackage{algorithmic}
\usepackage{natbib}
\usepackage{caption}
\usepackage{subcaption}
\usepackage{appendix}
\usepackage{amsthm}	

\graphicspath{{figures/}}
\title{Should All Temporal Difference Learning Use Emphasis?}

\author{
Xiang Gu \\
Department of Computer Science \\
Shanghai Jiao Tong University \\
\texttt{xiang7@ualberta.ca} \\
\And
Sina Ghiassian \\
Department of Computing Science \\
University of Alberta \\
\texttt{ghiassia@ualberta.ca} \\
\AND
Richard S. Sutton \\
Department of Computing Science \\
University of Alberta \\
\texttt{rsutton@ualberta.ca} \\
}

\begin{document}

\maketitle

\begin{abstract}
Emphatic Temporal Difference (ETD) learning has recently been proposed as a convergent off-policy learning method. ETD was proposed mainly to address convergence issues of conventional Temporal Difference (TD) learning under off-policy training but it is different from conventional TD learning even under on-policy training. A simple counterexample provided back in 2017 pointed to a potential class of problems where ETD converges but TD diverges. In this paper, we empirically show that ETD converges on a few other well-known on-policy experiments whereas TD either diverges or performs poorly. We also show that ETD outperforms TD on the mountain car prediction problem. Our results, together with a similar pattern observed under off-policy training in prior works, suggest that ETD might be a good substitute over conventional TD.
\end{abstract}


\acknowledgements{The authors gratefully acknowledge funding from Alberta Innovates--Technology Futures, the Natural Sciences and Engineering Research Council of Canada, and Google DeepMind.}

\startmain 

\section{Possible Advantages of Learning with Emphasis}
ETD was proposed by \citep{mahmood2015emphatic} \citep{sutton2016emphatic} as a stable and efficient method under off-policy training, and was later proven convergent \citep{yu2015convergence}. It utilizes the idea of emphasis -- re-weighting the states by emphasizing the update of some and de-emphasizing others -- to partially correct the updates to be under the on-policy distribution. With this restoration in the state distribution, existing results of convergence properties for the on-policy TD learning carries through to the off-policy case and it, therefore, makes ETD a stable learning algorithm even under off-policy training.

Surprisingly, ETD's on-policy version is also found to be significantly different from the conventional TD method. Yu's counterexample \citep{ghiassian2017first} highlights this difference by providing a two-state on-policy problem in which TD diverged but ETD converged to the correct solution.

A natural question to ask at this point is if on-policy ETD outperforms on-policy TD in all cases. We have seen evidences, the successful convergence of ETD on the Baird's counterexample (\citep{sutton2018reinforcement}) as well as several other experiments did in (\citep{ghiassian2017first}), that suggest ETD is a better algorithm than TD in the off-policy case. We do not know, however, whether a similar pattern carries through to the on-policy case. In this paper, we will thus focus on the on-policy case and test the performance of both methods empirically on several problems. We hope the results could provide empirical evidence for further investigation of ETD.

\section{On-Policy Emphatic TD($\lambda$)}
Consider the classic diagram of the agent-environment interaction in RL over a sequence of discrete time steps $t = 0, 1, 2, 3, \dots$. At each time step $t$, the agent receives some representation of the environment which we call state $S_t \in \mathcal{S}$ and based on that the agent chooses an action $A_t \in \mathcal{A}$ to take. The environment in return responds with a scalar number that we call reward $R_{t+1} \in \mathcal{R}$ and transitions to another state $S_{t+1} \in \mathcal{S}$. A state-dependent discounted factor $\gamma_t = \gamma(S_t)\in [0,1]$ is used. A policy $\pi$ is a mapping from states-action pairs to probabilities $\mathcal{S} \times \mathcal{A} \rightarrow [0,1]$ and we denote $\pi(a|s)$ as the probability of selecting action $a$ in state $s$.  State values $v_\pi(s)$ is then defined as the expected cumulative future reward if the agent starts at $s$ and follows policy $\pi$ thereafter $v_\pi(s) = \mathbb{E}_\pi[R_{t+1} + \gamma_{t+1} R_{t+2} + \gamma_{t+1}\gamma_{t+2}R_{t+3} + \dots | S_t = s]$.\\
We consider the situation where linear function approximation is used. In this case, states are represented as feature vectors $\mathbf{x}_t = \mathbf{x}(S_t) \in \mathbb{R}^d$ and state values are approximated by a linear function approximator $\hat{v}_\pi(S_t, \mathbf{w}) = \mathbf{w}_t^T \mathbf{x}_t$ where $\mathbf{w}_t \in \mathbb{R}^d$ is the weight vector at time $t$. We seek to learn a weight vector $\mathbf{w}$ such that $\hat{v}_\pi(s, \mathbf{w}) \approx v_\pi(s)$.  In fact, all actions can be selected by an alternative policy $b$. If $\pi = b$, then the training is called on-policy; Otherwise, it is called off-policy.

Emphatic TD$(\lambda)$ is the extension of ETD to eligibility traces -- a general mechanism that enables shifting and choosing from Monte-Carlo and TD method while maintaining the incremental update property. We consider the on-policy version of Emphatic TD($\lambda$) ($\rho_t = 1$, $ \forall t$), which is similar yet still significantly different to the conventional TD($\lambda)$ algorithm \footnote{When $\lambda = 1$ (no bootstrapping), this Emphatic TD($\lambda$) algorithm becomes identical to the conventional TD($\lambda$) algorithm.}. The algorithm can be completely described as:\\ 
\begin{eqnarray}
\mathbf{w}_{t+1} = \mathbf{w}_t + \alpha(R_{t+1} + \gamma_{t+1}\mathbf{w}_t^T\mathbf{x}_{t+1} - \mathbf{w}_t^T\mathbf{x}_t)\mathbf{e}_t \\
\mathbf{e}_t = \gamma_t\lambda_t\mathbf{e}_{t-1} + M_t\mathbf{x}_t,     \text{                with $\mathbf{e}_{-1} = \mathbf{0}$} \\
M_t = \lambda_t i(S_t) + (1-\lambda_t)F_t \\
F_t = \gamma_t F_{t-1} + i(S_t),     \text{                with $F_0=i(S_0)$}
\end{eqnarray}
where $\alpha > 0$ is the step size parameter, $\mathbf{e}_t \in \mathbb{R}^d$ is the eligibility trace at time step $t$, $\lambda_t = \lambda(S_t) \in [0,1]$ is the generalized state-dependent decay rate at time step $t$, $i(S_t) \in [0,\infty)$ is the user-specified interest on state $S_t$, $M_t$ is the emphasis of the update for $S_t$, and $F_t$ is an intermediate quantity (also referred to as the follow-on trace) from which we compute $M_t$.

In all experiments in this paper, we applied both TD($\lambda$) and Emphatic TD($\lambda$) with the interest for all states set to 1.

\section{Performance on Several Known On-Policy TD Counterexamples}
We begin our discussion on examining the performance of on-policy Emphatic TD($\lambda$) on several known counterexamples on which TD($\lambda$) is shown to either diverge or have a poor performance.

\begin{figure}[ht]
    \centering
    \includegraphics[scale=0.13]{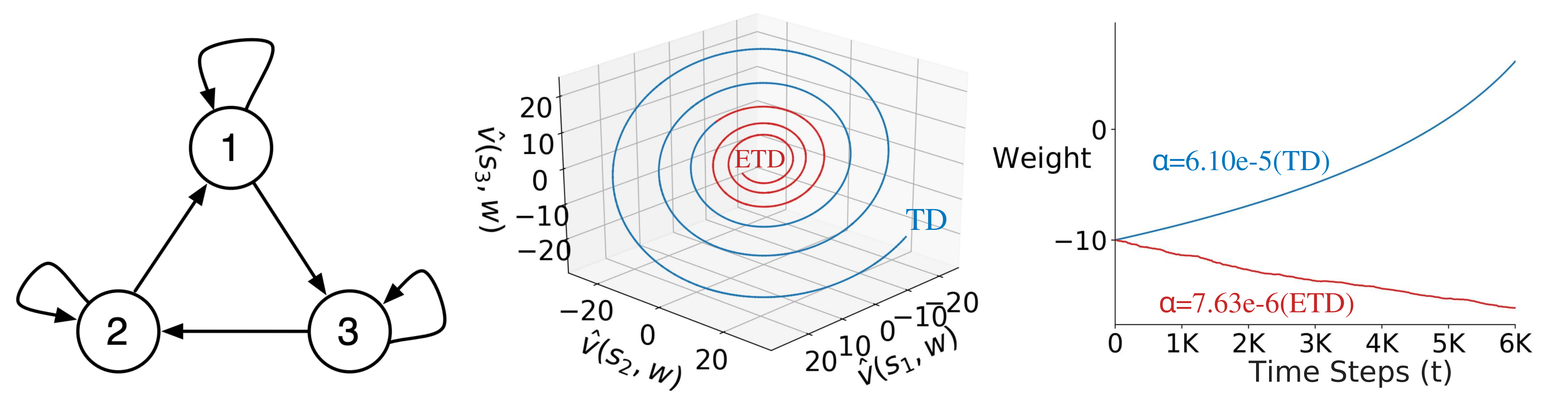}
    \caption{The MRP (left panel); Change of estimated state values (middle panel) and weights (right panel).}
    \label{fig:Spiral_Counterexample_MRP}
\end{figure}

The first counterexample is the 'spiral' counterexample introduced by \citep{tsitsiklis1997analysis} where on-policy TD(0) can diverge when a non-linear function approximator is used. Namely, consider a simple three-state Markov Reward Process (MRP) in figure \ref{fig:Spiral_Counterexample_MRP}. The probability of all transitions is 0.5 and the reward is 0 on all transitions. Discounted factor $\gamma \in (0, 1)$ is used. The task is to evaluate the state values for these three states, whose true values are $(0,0,0)^T$ \footnote{In this paper, $(x,y,z)$ represents a length-three row vector and thus $(x,y,z)^T$ represents a length-three column vector.}, with a non-linear function approximator $\hat{v}(s,w) \in \mathbb{R}^3$ parameterized by $w \in \mathbb{R}$. The explicitly form of $\hat{v}(s,w)$ can be expressed as
\begin{equation}
    \hat{v}(s,w) = e^{Aw}\hat{v}(s, 0),
\end{equation}
where A = 
$\begin{bmatrix}
1 + \epsilon & 0.5 & 1.5\\
1.5 & 1 + \epsilon & 0.5\\
0.5 & 1.5 & 1 + \epsilon
\end{bmatrix}$ 
and $(1,1,1)\hat{v}(s, 0) = 0$.

In our experiment, we chose $\gamma = 0.9$, $\epsilon = 0.05$, and $\hat{v}(s, 0) = (10, 10, -20)^T$. We applied both TD($\lambda$) and Emphatic TD($\lambda$) with constant $\lambda$ being 0 to the problem. The weight $w$ was initialized to -10 in both methods. Learning rates were picked to be such in the right panel in figure \ref{fig:Spiral_Counterexample_MRP}. The change of estimated values and weights were then plotted in figure \ref{fig:Spiral_Counterexample_MRP}.

One can see that the TD(0) method increases the parameter and the corresponding estimated state values spirals outward, whereas on-policy Emphatic TD(0) method drives the weights down and the estimated state values therefore spirals inward to the true state values $(0,0,0)^T$, which is achieved when $w$ goes down to negative infinity.

\begin{figure}[ht]
    \centering
    \includegraphics[width=\textwidth]{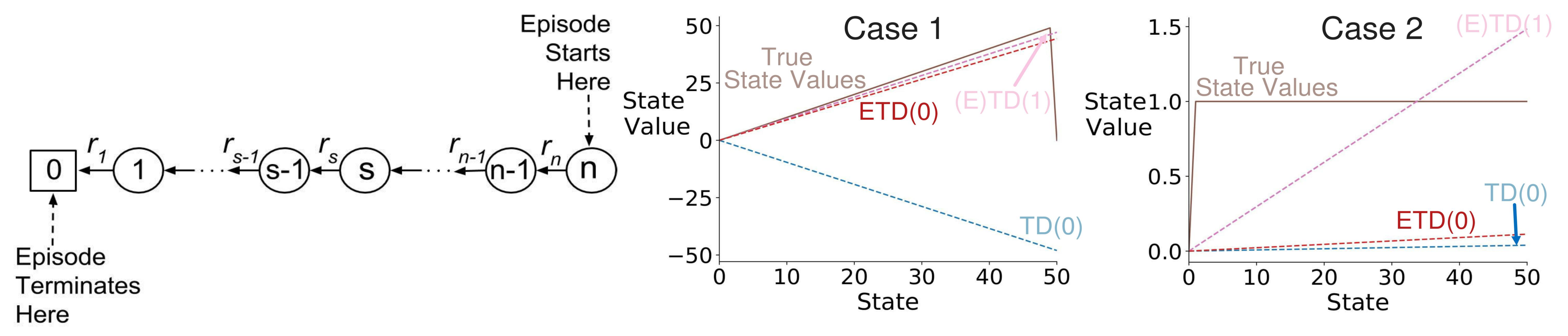}
    \caption{The MRP (left panel); Learned state values in case 1 (middle panel) and case 2 (right panel). The learning rates are omitted because you can reproduce the same results as long as your learning rate is sufficiently small.}
    \label{fig:Bertsekas_counterexample}
\end{figure}

The second counterexample is from \citep{bertsekas1995counterexample} where on-policy TD($\lambda$) is shown to become progressively poor as $\lambda$ switches from $1$ to $0$. As shown in figure \ref{fig:Bertsekas_counterexample}, consider a MRP with $n$ states. Each state $s \in \{1,2,3,...,n\}$ deterministically transitions to the previous state $s-1$ with reward $r_s$. The episode always starts at state $n$ and terminates after transitioning to state $0$ (the terminal state). A linear function approximaor $\hat{v}(s,w) = ws$ is used to approximate the state values ($w \in \mathbb{R}$).

We experimented TD($\lambda$) and Emphatic TD($\lambda$) with two constant $\lambda$'s (0 and 1) on two cases of the problem:
\begin{eqnarray*}
\text{case 1: } n = 50, r_n = -(n-1) \text{ and } r_s = 1 \text{, 
} \forall s \neq n \\
\text{case 2: } n = 50, r_1 = 1 \text{ and } r_s = 0 \text{, 
} \forall s \neq 1
\end{eqnarray*}
We then ran the algorithms until both methods converged and plotted the learned state values in figure \ref{fig:Bertsekas_counterexample}. 

In both cases, on-policy ETD(0) has a better performance than TD(0). In case one, the performance of TD($\lambda$) progressively degenerates when $\lambda$ switches from 1 to 0. In this designed case, the weight $w$ that TD(0) finds is almost an negation of that of TD(1). The on-policy ETD(0), however, does not experience this deterioration. It finds reasonably good solutions for both $\lambda$'s. In case two, ETD(0) also finds a better asymptotic solution than TD(0), albeit it is not very satisfactory either.

\begin{figure}[ht]
    \centering
    \includegraphics[scale=0.18]{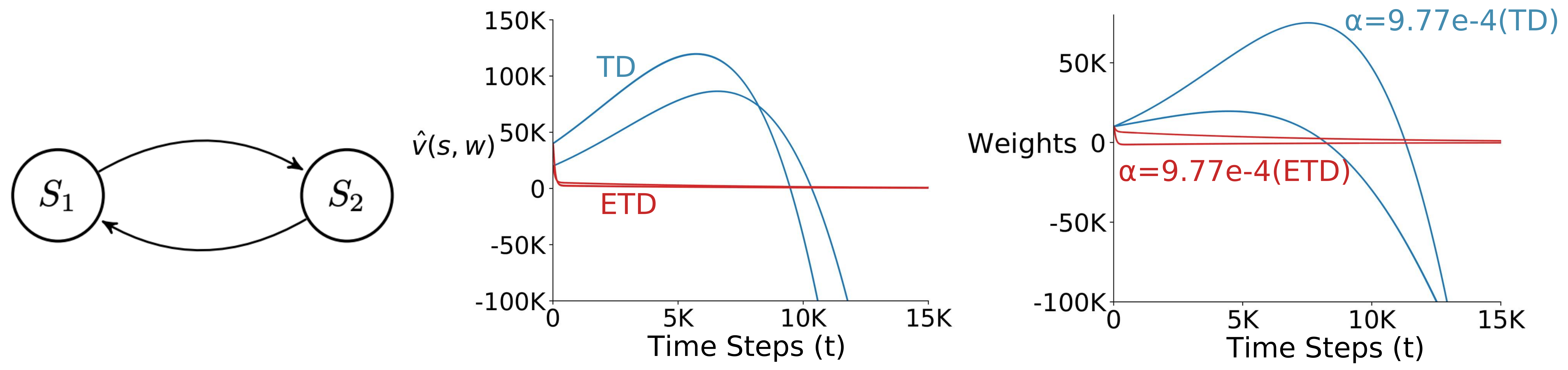}
    \caption{The MRP (left panel); Change of estimated state values (middle panel) and weights (right panel).}
    \label{fig:Yu_Counterexample}
\end{figure}

The final counterexample from \citep{ghiassian2017first} involves learning with variable $\lambda$'s. Consider a two-state MRP as shown in the left panel of figure \ref{fig:Yu_Counterexample}. One state deterministically transitions to the other with zero reward. This continuing task starts at one of the two states with equal probability. A linear function approximator $\hat{v}(s,w) = \mathbf{w}^T \mathbf{x}(s)$ is used where $\mathbf{x}(s_1) = (3,1)^T$ and $\mathbf{x}(s_2) = (1, 1)^T$. This special counterexample has state-dependent decay rate $\lambda(s_1) = 0$, $\lambda(s_2) = 1$ with a constant discounted factor $\gamma = 0.95$. Obviously, true state values $(0,0)^T$ is achieved when $\mathbf{w} = (0, 0)^T$. 

Again, we applied both TD($\lambda$) and Emphatic TD($\lambda$) to this problem. Our experiment initialized $\mathbf{w}$ to $\mathbf{w}_0 = (10^4, 10^4)^T$. We set the learning rate to be such indicated in the right panel of figure \ref{fig:Yu_Counterexample}. We plotted the change of the estimated state values and weights in \ref{fig:Yu_Counterexample}.

We can see that TD($\lambda$) in this counterexample diverges both in the weights and the estimated values. ETD($\lambda$), however, pushes the weights towards zero and the estimated values converge to the true values $(0,0)^T$ accordingly.

\section{Prediction: Fixed Policy Evaluation on Mountain Car}
In this section, we test our algorithms on a classic RL prediction problem -- a variation of Mountain Car problem introduced in \citep{sutton2018reinforcement}. The state consists of two variables -- position $x_t$ and velocity $\Dot{x}_t$. At each state, there are three actions -- full throttle forward (+1), full throttle reverse (-1) and zero throttle (0). The car moves according to a simplified physics with a forced bound: $-1.2 \leq x_{t+1} \leq 0.5$ and $-0.07 \leq \Dot{x}_{t+1} \leq 0.07$. In addition, when $x_{t+1}$ goes below the left bound, it stays at -1.2 but $\Dot{x}_{t+1}$ is reset to zero; When $x_{t+1}$ reaches the right bound, the goal is reached and the episode is terminated. The reward is -1 on all time steps before reaching the goal. Each episode starts from a random position $x_0 \in [-0.6, -0.4]$ and zero velocity.

\begin{figure}[ht]
    \centering
    \includegraphics[width=\textwidth]{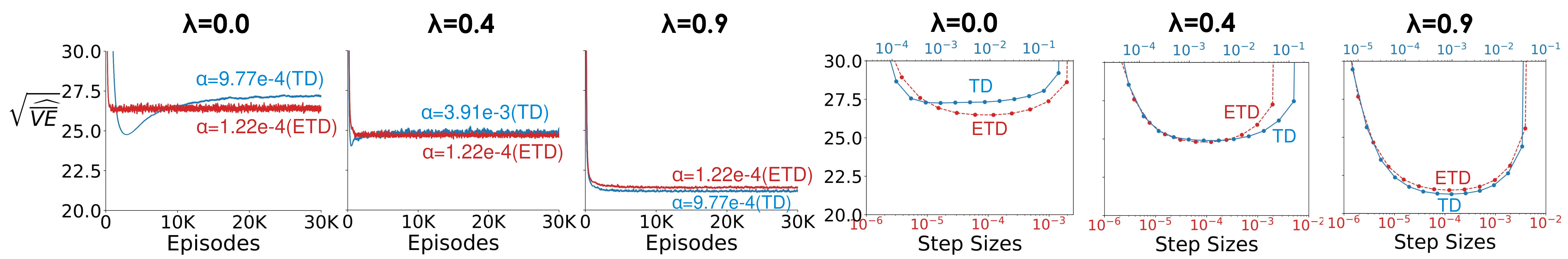}
    \caption{Results of Mountain Car prediction. The first three figures are the comparison of the best learning curve under three different $\lambda$'s using AUC criterion. I.e., the learning curve with the step size that gives the lowest estimated RMSE among all tested step sizes under the same $\lambda$. The rest three figures represent the comparison of TD($\lambda$) and ETD($\lambda$) methods for all tested step sizes under the same $\lambda$. Standard error are also shown at error bars but they are invisible due to their small sizes.}
    \label{fig:MountainCar}
\end{figure}

In our experiment, we evaluated a simple yet very effective policy $\pi$ that always selects the action that is in the same direction of the velocity $A_t = sign(\Dot{x}_t)$. We recorded the estimated root-mean-square error (RMSE) at each episode as the performance measure, which reflects the quality of our function approximator by computing the squared error between the estimated state values and the true state values, weighted according to how often each state is visited by following $\pi$:
\begin{equation}
    \sqrt{\widehat{\overline{VE}}(\mathbf{w})} = \sqrt{\frac{1}{|\mathcal{S}|}\sum_{s\in \mathcal{S}} i(s) (v_\pi(s, \mathbf{w}) - v_\pi(s))^2 } = \sqrt{\frac{1}{|\mathcal{S}|}\sum_{s\in \mathcal{S}} (v_\pi(s, \mathbf{w}) - v_\pi(s))^2 },
\end{equation}
where $\mathcal{S}$ contained 500 states gathered by simulating $\pi$ for 10,000,000 steps and randomly choosing 500 states from the last 5,000,000. We discarded the first half of the steps because the on-policy state distribution in episodic tasks is achieved in the limit and it might be different in the early stage of the simulation due to the randomness of the starting position. The true state values $v_\pi(s)$ were then computed and recorded by simulating $\pi$ from those 500 states individually till the termination of the episode. \\
We tested our algorithms on this environment with three different $\lambda$'s -- $0.0$, $0.4$, and $0.9$. For each $\lambda$, we tested on a wide range of different step size $\alpha$'s. For each $\lambda$ and $\alpha$ combination, we repeated the experiment for 30 trials to average out the randomness where each trial lasted for 50,000 episodes.\\
After getting all the data, we chose to visualize them using the Area Under the Curve (AUC) criterion: For each trial, we averaged over the 50,000 estimated RMSEs recorded in this trial to compute the mean RMSEs across all episodes in this trial; We then repeated the same process for all the 30 different trials, all of which are of the same parameter setting, and ended up with 30 mean RMSEs. Finally, we averaged over these 30 mean RMSEs to compute the final estimated RMSE under its corresponding method and parameters. At the same time, we also computed the standard deviation of those 30 mean RMSEs as the standard error of the final estimated RMSE. Results are shown in figure \ref{fig:MountainCar}.

We can see that Emphatic TD($\lambda$) finds a fixed point $w$ that has better asymptotic performance as well as faster learning rate when $\lambda = 0.0$. As $\lambda$ goes up toward 1, the performance gap begins to decrease and both methods start to converge to the same performance when $\lambda$ reaches 1. One interesting observation is the 'bump' of the TD(0) method -- the function approximator once learned a better weights but then it gradually degenerated and converged to a weights that is worse -- when $\lambda=0$. This bump gradually disappears as $\lambda$ increases toward 1. Emphatic TD($\lambda$) did not experience this bounce. Instead, it updates the weights so that the performance measure monotonically decreases. We do not know why this occurs to TD($\lambda$) method at the moment and a better understanding of the algorithm is required to explain it.

\section{Conclusion and Future Work}

In the paper, we first showed the on-policy Emphatic TD($\lambda$) algorithm performs well on several known counterexamples compared to the TD($\lambda$) algorithm. In particular, the result on the Tsitsiklis and Von Roy's counterexample is encouraging. It opens the possibility of further investigating on on-policy Emphatic TD($\lambda$) methods with non-linear function approximator, and one conjuncture can therefore be proposed that on-policy Emphatic TD($\lambda$) methods are convergent with any non-linear function approximator.   

For the prediction experiment we did on Mountain Car environment, Emphatic TD($\lambda$) also gives better asymptotic performance as well as learning speed. 
We also tried to extend this to control problem by extending equation 1 to action values and utilizing the generalized policy iteration (GPI) framework. The results, however, have several potential issues and the there is so far no insight we can gain from the control experiment, so we will reserve our judgement on the control case.

This experiment and the results on the Bertsekas's counterexample, together with several other results in \citep{ghiassian2017first} and \citep{ghiassian2018online}, showed a consistent pattern of favoring Emphatic TD($\lambda$) over conventional TD($\lambda$) in the on-policy case. We did not strive to find a counterexample to challenge it. Nevertheless, they are strong suggestive evidence for another conjecture that Emphatic TD($\lambda$) finds a better fixed point asymptotically as well as faster learning speed, which is also our first attempt to answer the question introduced in the first section.

An interesting future work direction is to vary interest for different states as described in the original ETD paper. For all of our experiments, we used uniform interests on all the states (i.e., $i(s) = 1$, $\forall s$). In fact, the introduction of interest in ETD methods offers an additional degree of freedom to specify the 'importance' of each state, rather than relying on the state visitation frequency as a sole weighting. There are conceivable scenarios where one wants to evaluate states that are less often visited more accurately. 

\bibliography{references}

\begin{thebibliography}{8}
\providecommand{\natexlab}[1]{#1}
\providecommand{\url}[1]{\texttt{#1}}
\expandafter\ifx\csname urlstyle\endcsname\relax
  \providecommand{\doi}[1]{doi: #1}\else
  \providecommand{\doi}{doi: \begingroup \urlstyle{rm}\Url}\fi

\bibitem[Mahmood et~al.(2015)Mahmood, Yu, White, and
  Sutton]{mahmood2015emphatic}
A~Rupam Mahmood, Huizhen Yu, Martha White, and Richard~S Sutton.
\newblock Emphatic temporal-difference learning.
\newblock \emph{arXiv preprint arXiv:1507.01569}, 2015.

\bibitem[Sutton et~al.(2016)Sutton, Mahmood, and White]{sutton2016emphatic}
Richard~S Sutton, A~Rupam Mahmood, and Martha White.
\newblock An emphatic approach to the problem of off-policy temporal-difference
  learning.
\newblock \emph{The Journal of Machine Learning Research}, 17\penalty0
  (1):\penalty0 2603--2631, 2016.

\bibitem[Yu(2015)]{yu2015convergence}
Huizhen Yu.
\newblock On convergence of emphatic temporal-difference learning.
\newblock In \emph{Conference on Learning Theory}, pages 1724--1751, 2015.

\bibitem[Ghiassian et~al.(2017)Ghiassian, Rafiee, and
  Sutton]{ghiassian2017first}
Sina Ghiassian, Banafsheh Rafiee, and Richard~S Sutton.
\newblock A first empirical study of emphatic temporal difference learning.
\newblock \emph{arXiv preprint arXiv:1705.04185}, 2017.

\bibitem[Sutton and Barto(2018)]{sutton2018reinforcement}
Richard~S Sutton and Andrew~G Barto.
\newblock \emph{Reinforcement learning: An introduction}.
\newblock MIT press, 2018.

\bibitem[Tsitsiklis and Van~Roy(1997)]{tsitsiklis1997analysis}
John~N Tsitsiklis and Benjamin Van~Roy.
\newblock An analysis of temporal-difference learning with function
  approximation.
\newblock \emph{IEEE TRANSACTIONS ON AUTOMATIC CONTROL}, 42\penalty0 (5), 1997.

\bibitem[Bertsekas(1995)]{bertsekas1995counterexample}
Dimitri~P Bertsekas.
\newblock A counterexample to temporal differences learning.
\newblock \emph{Neural computation}, 7\penalty0 (2):\penalty0 270--279, 1995.

\bibitem[Ghiassian et~al.(2018)Ghiassian, Patterson, White, Sutton, and
  White]{ghiassian2018online}
Sina Ghiassian, Andrew Patterson, Martha White, Richard~S Sutton, and Adam
  White.
\newblock Online off-policy prediction.
\newblock \emph{arXiv preprint arXiv:1811.02597}, 2018.

\end{thebibliography}

\end{document}